# Distributed Artificial Intelligence as a Means to Achieve Self-X-Functions for Increasing Resilience: the First Steps

Oxana Shamilyan[1,2], Ievgen Kabin[1], Zoya Dyka[1] and Peter Langendoerfer[1,2]
[1] IHP – Leibniz-Institut für innovative Mikroelektronik, Frankfurt (Oder), Germany
[2] BTU Cottbus-Senftenberg, Cottbus, Germany
{shamilyan, kabin, dyka, langendoerfer}@ihp-microelectronics.com

*Abstract*—**Using sensors as a means to achieve self-awareness and artificial intelligence for decision-making, may be a way to make complex systems self-adaptive, autonomous and resilient. Investigating the combination of distributed artificial intelligence methods and bio-inspired robotics can provide results that will be helpful for implementing autonomy of such robots and other complex systems. In this paper, we describe Distributed Artificial Intelligence application area, the most common examples of continuum robots and provide a description of our first steps towards implementing distributed control.**

## I. Introduction

Nowadays resilience is considered being one of the main properties to provide stability and security for cyber-physical systems (CPS) or cyber-physical systems of systems (CPSoS). There are some examples of such CPS(oS): Smart home, Industry 4.0, Internet-of-Thing (IoT), E-health and many others. Resilience can be defined as a set of three components/features: reliability, security and autonomy. Each of these components can be reached by means of different factors. For implementing security requirements such as communication confidentiality, (mutual) authentication of the communication parts, data integrity, services availability etc. cryptographic algorithms are used. An important aspect is the resistance of cryptographic implementations against tamper and side-channel analysis attacks. Especially dangerous are horizontal side-channel analysis attacks against digital signature implementations [1]-[3] which successfully can cause the loss of identity. A usual means to increase the resistance against fault injection attacks and reach reliability is the redundancy. The main means to reach system autonomy can be the application of artificial intelligence (AI) approaches. System autonomy is a set of many "self-functions", such as self-monitoring, self-awareness, self-restoring, self-reconfigurability, self-adaptability, etc. denoted often as "Self-X-functions". Implementing the Self-X-functions means that the system is able to learn and to make decisions by its own, which leads to the necessity of using AI methods. Improving all of these features is in particular going to increase the resilience of a system in general.

Distribution of tasks among the system nodes or/and according to a specific time schedule can reduce the energy consumption of the system. Combined with redundancy, this strategy can increase the resistance of the system against natural faults, manipulations, and even against some side-channel analysis attacks, increasing the uptime of the system. Thus, existing distributed control techniques, such as distributed decision-making, can make the system more resilient. The necessity of using this approach appears first of all in large and complex systems, where each node holds only limited information, and the cooperation between nodes is critical for the system's performance [4]. There are some cases, in which distributed decision-making can be useful. E.g. if fast data processing is required due to the rapid changes of a system state, i.e. for a system of moving cars, which transmit information about their current state among each other. Another aspect is that data transmission between nodes is an energy-consuming process and can eventually be omitted if techniques of distributed control are applied to allow data processing (at least partly) directly by a node instead of sending data.

More application examples of distributed control can be found in such areas as Wireless Sensor Networks (WSNs), Internet-of-Thing (IoT), Robotics, etc. In this paper we are focusing on discussing how to implement distributed decision-making methods in the Robotics area. There are different types of robots applied for industrial, medical, research, and other tasks. There are discrete and continuum robots. Discrete robots usually consist of rigid parts connected with each other by joint links. Continuum robots have much less joint links than discrete robots or they don´t have them at all. They have an infinite number of degrees of freedom that makes them more flexible and sometimes they are partially or completely made of soft materials (soft-robots) [5].

Most of the robots have a centralized control, are not autonomous and cannot perform the task without human supervision. [6] includes a description of the levels of autonomy for medical robots. There are five different levels from 0 to 5, where robots with level 0 have no autonomy and robots with level 5 are fully autonomous. Autonomy in robots can be described by robot abilities of detecting and tracking a target, environmental assessment, task planning. These properties can be achieved by applying Distributed Artificial Intelligence (DAI) methods.

In this paper we are discussing two science areas – distributed artificial intelligence and continuum robotics – and present our first steps of implementing control distribution to continuum robots as a result of application of DAI methods.

The rest of this paper is structured as follows. Section II describes DAI application areas. Section III gives a short overview of the most common types of continuum robots and principles of their work. In Section IV we describe some details about our continuum robot prototype that we implemented and discuss its advantages for our future work. A short conclusion is given in section V.

## II. DAI AND THEIR APPLICATIONS

Distributed Artificial Intelligence (DAI) [7] is a class of technologies in which the solving of a problem is distributed between "agents". To solve the problem a set of tasks can be defined. These tasks can be divided into more or less independent subtasks. Each subtask is assigned to an agent which starts to solve it. Communication between agents depends on the system configuration, but usually they communicate with each other by sending updates about their current state.

DAI can enhance the system robustness that increases the resilience of the whole system [8]. The main goal is to make a system self-adaptive and self-aware with the ability to change its behaviour by itself according to the current state of its environment. Agents can change their algorithm to solve their subtasks according to updates received from neighbours. The result is a system that can make decisions on its own about what to do, in which order, and how exactly, i.e. it is an intelligent system.

DAI approaches can be applied for systems in which it is possible to distinguish independent agents and to distribute control among them. DAI can be used in such areas as the Internet of Things (IoT) and Wireless Sensor Networks (WSNs). [9] provides a big survey containing more than thirty researches made in IoT. Authors are classified DAI approaches into five different groups: cloud-computing, mist-computing, service-oriented computing, distributed-ledger-technology, and hybrid technology. Also the survey presents a taxonomy of DAI approaches in IoT. [10] contains information about application of DAI in WSNs. The usage of DAI techniques provides many advantages for systems due to increasing adaptability, flexibility, security and scalability. Implementation of DAI in areas such as IoT and WSNs can be considered as an adaptation of a human behavior into technical solutions. Human organizations are good examples of distributed intelligence, e.g. surgery and military teams, aircraft crew. These teams are very productive due to the distribution of a responsibility for solving the tasks [11].

Gaining knowledge from a biology area is a modern way to develop a robot [5]. AI methods are often used here as well. They are used as means of increasing robots resilient properties. In our previous paper [12] we consider the octopuses as an interesting example of a biological system.

Octopuses are one of the most popular cephalopods. Researchers are interested in their morphology, body arrangement and expanded cognition. Octopuses' body is bilaterally symmetrical, with eight arms and without any rigid parts. The nervous system is the most interesting part of the octopus's body. For adult individuals the number of nerve cells can reach the point of 300-500 million. The complexity of nervous system and cognition makes octopuses very intelligent [13], [14]. The octopus intelligence is not centralized and divided into two groups: the central nervous system (CNS) and the peripheral nervous (PNS) system [13], [15], [16]. The central nervous system is the first who makes decisions, initializes the nerve impulse and sends it to PNS. Nevertheless, it doesn´t initialize impulses for every decision and movement. Most decisions the PNS makes by itself. This specificity of arrangement fits well under the definition of distributed artificial intelligence and can be used as the prototype for system control architecture.

Octopuses and especially octopus arms are also interesting for robotics researchers. They are used as inspiration to develop robots such as robotic arms. An octopus-like robotic arm is a hyper-redundant robot manipulator that is high flexible and has many cases of usage, such as surgery, explore and rescue operations. These robots are also called continuum robots. Their shape and structure are very elastic with infinity number of joints and degrees of freedom [5]. More details about existing technical octopus-inspired solutions can be found in [12].

## III. CONTINUUM ROBOTS

Continuum robots is a novel concept of robot design. Unlike the usual rigid robots, they possess higher flexibility, dexterity, infinite degrees of freedom and lower weight. These properties are achieved mainly due to the absence of rigid links. Continuum robots benefits make them on the one hand more resilient, but on the other hand more nonlinear and thereby harder to control [17]. For continuum robots' control different mathematical models, which define movements of robots, their accuracy and dexterous, are used.

Understanding of a mathematical model, used for the description of robot movements, can provide useful information for developers. For example, it can help to understand, which particular movements this model can describe (bending, fetching, etc.) or which environmental forces developers should take into account, to obtain more precise movements. Using this knowledge the model can be modified or improved according to the developer's needs, therefore the resulting robot will obtain more accuracy and dexterity.

Further in this section we provide a short overview of some mathematical models, which are widely used by researchers and developers, as well as a comparison of robots, which are based on the described mathematical models.

*A. Overview of mathematical models*

The existing mathematical models can be distinguished as follows: piecewise constant curvature (PCC) [18], continuous Cosserat model [19] and 3D dynamic model based on Mode Shape Functions (MSF) [20].

The PCC approach is used by developers of continuum robots in most of the cases due to its simplification of the kinematic modelling. In this approach the soft prototype is divided into a finite set of circular arcs. It is easy to implement, however the PCC approach does not suit for each situation, especially it is not valid under non-negligible external loads including gravity [21].

The continuum Cosserat approach is used for models with infinite degrees of freedom. Robots are represented by a stack of infinitesimal small solids parts. Initially this concept was used for hyper-redundant robots and recently has found its application in soft robotics locomotion. In opposite to PCC, Cosserat can be implemented in models with external loads. It allows torsion and shears to be calculated along with curvature and elongation. Cosserat approach has many variations [19], [21]-[24], nevertheless the model basics are always the same.

3D dynamic model produces correct results for pure extension/contraction, and bending. The use of 3D dynamic model based on Mode shape functions simplifies the model, and thereby allows considering complex effects such as drag, lift, gravity, distributed mass, angular moments and linear forces.

*B. Comparison of continuum robots*

The field of continuum robots is a new and actively improving field. Researchers all over the world start to design their own continuum robots. They combine different techniques and approaches to achieve better results in comparison with rigid robots. There are continuum robots inspired by octopus arms, snakes, chameleon tongues, vines and plants. They can be single- or multi-segmented and can be made of silicone, aluminum, polycarbonate, shape memory alloy, etc.

Continuum robots have a few significant benefits compared to rigid robots. For example navigation in highly unstructured, unexplored or narrow environments. Their potential application areas are for example inspection tasks, handling of unstructured objects, and minimally invasive surgery.

During our research we found both, single researcher groups [25]-[29] and research laboratories [30]-[33]. They both provide useful information about continuum robots development, however we decided to focus on the laboratories, because they publish papers more often, with faster progress. For us it was easier and faster to understand the current state of the art in this field of research. The chosen laboratories work with the most common types of continuum robots that are: soft robots, tendon driven robots, and concentric tube robots. There are laboratories specialized in a certain type of robots:

- Soft Robotics Laboratory, The Biorobotics Institute, Scuola Superiore Sant'Anna, Pisa, Italy [30];
- The Continuum Robotics Laboratory (CRL), Toronto, Canada [31];
- Robotics and Medical Engineering (RoME) Lab, Chicago, the USA [32].

Table I provides an overview of three types of continuum robots and their essential parameters [34]-[36]. They were chosen because the laboratories were mostly focused on these types of continuum robots and provided a sufficient progress in their study and development in the last 10 years [34]-[39].

A silicone soft-robotic arm [34] was created based on the parameters and behaviour of real octopus arms. Movements of a real octopus were recorded and interpreted in a digital form using special markers. In addition, a computer simulation based on the Cosserat model was created and trained by machine learning. The main goal of this research is to achieve the most precise movements of a computer model.

A tendon-driven continuum robot developed by The Continuum Robotics Laboratory [35] is aimed mainly to perform a minimally invasive surgery. The robot is very small in diameter (approx. 8 mm) and as the first robot has a bio-inspired background. It´s based on [40] that is also one of the interpretations of Cosserat theory. The robot has manual control and its movements depend on tendon tension. It can perform bending, elongation, contraction movements.

A robot developed by Robotics and Medical Engineering Lab [36] also has computer simulation. Its movements are described by a 3D dynamic model, which is able to calculate bending, elongation, contraction, object inspection and fetching. This model is trained by machine learning to plan a path, avoid obstacles, inspect and handle objects of different shapes.

Even though some of the described robots are represented by computer models and trained by machine learning which makes them autonomous, we did not find any information about distributed control based on AI approaches. That means that the robot moves as a single system without any distribution within it. We also did not find any information about robots interaction, for example, if two or more robots implement one task together, helping each other.

IV. OUR FIRST EXPERIMENTS WITH A CONTINUUM ROBOT

Distributed AI methods are a promising approach to increase autonomy in a developed system. Though we did not find any proofs of implementation of control distribution, in [41] a real-time decision-making mechanism was mentioned as a means of increasing autonomy in future works. Continuum robots and computer-simulated models, described in the previous section, were built to achieve more precise movements, while our goal is to achieve distributed control. For this purpose, we need a suitable model of a continuum robot that allows us to concentrate on the mechanisms of the control distribution as well as on the decision-making mechanisms. Moreover, the model has to be able to learn the movements. Thus, we need a physical model – a prototype – for our experiments in future.

TABLE I. OVERVIEW OF THE MOST COMMON TYPES OF CONTINUUM ROBOTS

| Name of laboratory | Type of robot | Math model | Type of actuation | Workspace | Computer simulation | Implemented movements | Type of control | Field(s) of application |
|---|---|---|---|---|---|---|---|---|
| Soft Robotics Laboratory | Silicone soft-robotic arm driven by cables [34] | Cosserat model | Embedded cables | Underwater | Yes | Bending, reaching and fetching | Trained computer simulation model | Learning and practical field. Solution shows applicability of machine learning tools for soft robotic applications. |
| The Continuum Robotics Laboratory | Tendon-driven continuum robot with extensible sections [35] | Cosserat model | Tendons driven | Air | Yes | Bending, elongation, contraction | Manual control | Minimally invasive surgery |
| Robotics and Medical Engineering Lab | Biometric multi-section continuum arm [36] | 3D dynamic model | Hydraulic muscle actuator | Underwater | Yes | Bending, elongation, contraction, object inspection and fetching | Trained computer simulation model | Minimally invasive surgery, inspection of objects with irregular shapes, object handling |

For this reason we selected a tendon-driven continuum robot that represents an octopus arm. The initial robot design has been taken from the "Hackday" web-page [42]. It has been presented as a self-made project by Joshua Vasquez. More information about this solution can be found in [43], [44]. We selected this design as a good example of a robot with a detailed description that can be easily built. Here we identified the opportunity to change the control strategy from manual to distributed.

*A. Implemented prototype*

The prototype that we implemented corresponding to the description [44] has two stages controlled separately by two manual controllers. Each controller has pulleys, which can control tendons tension, thereby moving the tentacle. Fig. 1 shows the prototype we realized.

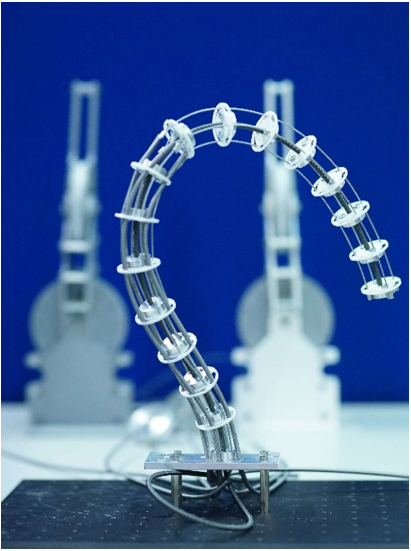

Figure 1.  Continuum tendon-driven robot prototype

A very important feature of this design is so called back-drivability. According to the [43] the term "back-drivability" means that, we can manipulate the tentacle by hands, and those movements will fed back up the cables to then drive the controller. In other words, it gives us an opportunity not only to move the tentacle by moving controllers, but also to move controllers by affecting tentacle stages. This prototype allows to implement our experiments with the distribution of the control of the robot.

*B. Discussion and future work*

During building our robot prototype, we faced a few challenges. We have printed all parts with a 3D printer using Raise 3D PLA Filament. It is a kind of industrial thermoplastic that is non-toxic, bio-based and 100% bio-degradable, but at the same time is relative fragile. Due to this fact, parts were not strong enough to handle pressure, hits and other forces and they started to break. We decided to change the plastic spine circles into metal ones. This made the tentacle part more strong and robust to external forces. Manual controllers with a system of pulleys, which we also printed using the 3D-printer, did not possess enough robustness to create sufficient tension of tendons. Our decision was to replace plastic controllers with a system of step-motors and joysticks connected to each other. Each pair of a step-motor and a joystick controls one separate stage of the tentacle. The system is controlled by Arduino microcontrollers.

After the automatization, AI methods can be used to train our prototype to implement some basic movements. Thus, the automatization of the control is the first step allowing our future experiments.

With the goal to show how control distribution and decision-making mechanisms work in real life, we are going to use two robots, which should complete one task by working together. Thereby, they represent a multi agents system, where each robot is one separate agent. Here we can also discuss system redundancy, e.g., when one of the two robots is unavailable, damaged or (partially) out of order. In this case, the working robot should be able to solve at least selected tasks completely. The selection of the tasks or subtasks depends on the still available functionality of the 2nd robot – both robots must decide to redistribute the steps of their task.

The goal of our scenario is to demonstrate that two robots can work as one team, helping each other to complete a task, hence, we can prove, self-organization, self-adaptability and control distribution.

## V. CONCLUSION

During our research we surveyed the area of bio-inspired continuum robots. In particular, we were looking for the way these robots are controlled and the presence of autonomy there. Continuum robotics is a fast improving area with many contributors. We identified three different laboratories with good results and impressing advances over the last years and provided an overview of robots they are developing. To the best of our knowledge there is no mention of applying DAI methods. Based on our research we choose a particular robot design – a tendon-driven continuum robot – as the best candidate to achieve our goal of implementing control distribution and distributed decision-making mechanisms. We implemented our prototype and discussed its advantages as well as our further steps for the future work.